\documentclass[10pt,letterpaper]{article} %

\usepackage[preprint,twocolumn,noalgo]{labpreprint}

\IfFileExists{algorithm.sty}{\usepackage{algorithm}}{%
  \usepackage{float}%
  \newfloat{algorithm}{tbp}{loa}%
  \floatname{algorithm}{Algorithm}%
}
\IfFileExists{algorithmic.sty}{\usepackage{algorithmic}}{}

\usepackage{multirow} 
\usepackage{newfloat}
\usepackage{listings}
\DeclareCaptionStyle{ruled}{labelfont=normalfont,labelsep=colon,strut=off} %
\floatstyle{ruled}
\newfloat{listing}{tb}{lst}{}
\floatname{listing}{Listing}

\newcommand{\monkey}{\raisebox{-0.18em}{\includegraphics[height=1.0em]{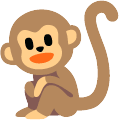}}}

\title{\monkey \ MACaQ: Fast, \underline{M}odel \underline{A}gnostic \underline{Ca}librated \underline{Q}uantization for Low-Precision Weights and Activations}

\author{\mbox{Lorenz K. Muller} \quad
        \mbox{Philippe Bich} \quad
        \mbox{Chiara Boretti} \quad
        \mbox{Hyun-Min Chang} \quad
        \mbox{Jiawei Zhuang} \quad
        \mbox{Lukas Cavigelli}}

\affiliation{Computing Systems Lab, Huawei Technologies, Switzerland.}

\laborg{Huawei}
\labname{Computing Systems Lab}
\makeatletter
\def\lp@logo{\includegraphics[height=1.15cm]{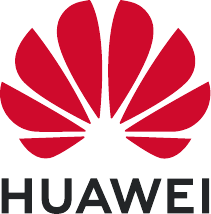}}
\makeatother
\date{}
\begin{document}
\maketitle

\begin{abstract}

Weight-activation quantization can reduce the memory and compute
cost of neural-network inference, but recent methods often require expensive
optimization and architecture-specific propagation of basis transforms across
layers. We introduce MACaQ, an MXFP4 W4A4 post-training quantization
method that treats every linear layer independently. MACaQ applies a
per-channel scale and a block-Hadamard transform online to each layer's
activations, while absorbing the corresponding inverse transforms into its
weights. Retaining these transforms online avoids cross-layer dependencies and
enables a simple streaming implementation of the quantization process,
in which only one layer must reside
in memory at a time. To improve over round-to-nearest quantization, we introduce
Flip-Refinement, a Hessian-aware coordinate-descent procedure that greedily
moves individual FP4 codes to adjacent representable values. Despite executing
the activation transforms at inference time, MACaQ adds only $2.3\%$
end-to-end overhead over pure MXFP4 on Qwen3-32B in vLLM. MACaQ reaches lower perplexity than DuQuant++ and MR-GPTQ, and already matches their best perplexity on Llama-3-8B in 11× less quantization time. We further apply MACaQ without
architecture-specific modifications to an image diffusion model, illustrating
the generality of its layer-local design and show quantization of a 122B parameter 
model in 64 minutes, using only 32.5GB GPU RAM. Code is available at \url{https://github.com/huawei-csl/MACaQ}.

\end{abstract}

\section{Introduction}
\label{sec:introduction}

Perhaps the central idea in recent neural network quantization strategies is the
change of basis. To compute the matrix product between an input and a weight
matrix, we may transform the input by a matrix $P$ and apply the
corresponding inverse transform to the weights, leaving the product unchanged.
If $P$ is chosen wisely, the transformed inputs and weights can be
substantially easier to quantize.

\begin{figure}
    \centering
    \includegraphics[width=\linewidth]{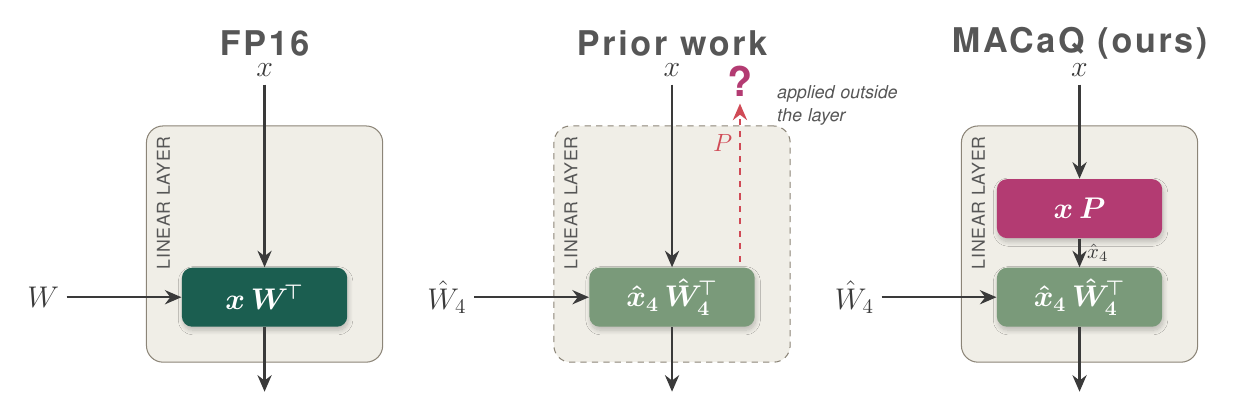}
    \caption{Prior work propagates the activation-side change of basis
    (labeled $P$) outside the linear layer to avoid runtime overhead. 
    This incurs an engineering and quantization-time overhead, 
    where for each layer a suitable absorption target must be found. 
    MACaQ instead applies a lightweight transform online, preserving complete
    layer locality at negligible (ca. 2\%) run-time overhead for sufficiently large layers.}
    \label{fig:macaq-overview}
\end{figure}

Two types of transform are widely used to ease quantization. First, activations
and weights may be rotated into a different basis using an orthogonal matrix
$M_{\mathrm{rot}}$. The Hadamard transform is especially popular
because it can be applied in $O(N\log N)$ operations and tends to produce
approximately Gaussian-distributed outputs \cite{quip}. Second, individual channels
may be rescaled using a diagonal matrix $M_{\mathrm{diag}}$
\cite{smoothquant, AWQ, SINQ}, thereby suppressing outlying channels and
balancing the quantization difficulty of activations and weights.

One potential downside of such a change of basis is that, although the
weight-side transform can be absorbed into the weights at quantization time,
the corresponding input transform ostensibly needs to be applied online. Prior
work often avoids this cost by propagating the input transform through the
network and absorbing it into preceding layers. This approach is widely
used because it can eliminate the explicit runtime transform
\cite{quarotb}. However, it also introduces dependencies between
layers: applying the transform to one linear layer may require modifying its
normalization parameters, preceding projections, residual branches, or other
connected operators.

In this paper, we argue for tolerating the overhead introduced by online
transforms because keeping them online provides a different benefit:
\emph{simplicity through layer locality}. MACaQ confines every transform and
quantization decision to a single linear layer. No neighboring layer is
modified to compensate for its quantization, and no basis transform needs to be
propagated through the network. Eliminating these inter-layer dependencies
allows each linear layer to be calibrated, transformed, and quantized
independently. This simplifies application to new architectures, enables layers
to be loaded and quantized one at a time, and leads to substantially faster
quantization. Furthermore, \emph{all} linear layers can benefit from online basis transforms,
where in some architectures not all layers can have absorbed basis transforms.

The cost of this design choice is small. MACaQ uses only a per-channel scale and
a block-Hadamard transform online. Their cost grows with the input dimension,
whereas the cost of the subsequent matrix multiplication also grows with the
output dimension. Consequently, their relative overhead diminishes for larger
layers. We measure an end-to-end decode overhead of $2.3\%$ for Qwen3-32B in
vLLM, while obtaining a better quality-quantization-time trade-off than prior
MXFP4 methods.

\begin{figure}[t]
\centering
\includegraphics[width=0.92\linewidth]{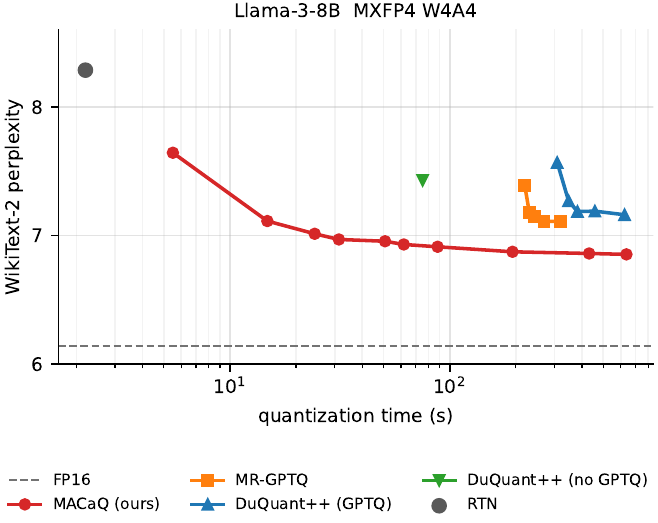}
\caption{Quantization time vs.\ WikiText-2 perplexity on Llama-3-8B (OCP MXFP4 W4A4),
all measured on the same hardware. MACaQ is run with different number of 
flips and calibration tokens, the other methods where possible scale by calibration tokens.
MACaQ matches the best quality of these strong baselines in $11\times$ less time.}
\label{fig:pareto}
\end{figure}

The main contributions of this paper are:
\begin{itemize}
    \item We propose \textbf{MACaQ}, a strictly per-linear-layer pipeline for W4A4 MXFP4
    quantization.
    MACaQ achieves higher output quality than
    state-of-the-art MXFP4 methods while requiring up to $11\times$ less
    quantization time, see Fig.~\ref{fig:pareto}.

    \item We show that retaining lightweight basis transforms \emph{online} incurs
    only a small inference overhead while \textbf{eliminating cross-layer dependencies}.
    This simplifies engineering, see Fig.~\ref{fig:macaq-overview}, enables layer-at-a-time
    quantization with low memory requirements, and allows the same method to
    work on widely different model architectures.

    \item We introduce \textbf{Flip-Refinement}, a fast Hessian-aware coordinate-descent
    method that improves an initial round-to-nearest quantization assignment by
    moving individual FP4 codes to adjacent representable values.

    \item We provide efficient \textbf{Triton kernels} for MACaQ-quantized MXFP4
    models in vLLM, achieving up to $3\times$ end-to-end speedup over FP16 while
    limiting the overhead of online transforms to $2.3\%$ on
    Qwen3-32B.
\end{itemize}

\section{Method}

\subsection{Problem Formulation}

We consider weight and activation quantization to MXFP4, with the matrix
multiplications themselves executed in MXFP4. An MXFP4 tensor is divided into
blocks of $B=32$ elements. Each block shares an E8M0 scale, represented by eight
exponent bits and no mantissa bits, while each element has an individual
four-bit FP4 E2M1 code consisting of a sign bit, two exponent bits, and one
mantissa bit \cite{rouhani2023microscaling}. %

Consider a linear layer with weight matrix
$W\in\mathbb{R}^{m\times n}$ and a calibration set of layer inputs
$X\in\mathbb{R}^{N\times n}$. Similar to GPTQ \cite{GPTQ}, we seek to minimize
the layer's output reconstruction error:
\begin{equation}
\label{Eq:HessianTarget}
\min_{\widehat W}\ 
\left\lVert XW^\top-X\widehat W^\top\right\rVert_F^2
=
\sum_{i=1}^{m} r_i H r_i^\top,
\end{equation}
where
\begin{equation}
H=X^\top X,
\qquad
r_i=w_i-\widehat w_i.
\end{equation}
Here, $w_i$ is the $i$-th row of $W$, $\widehat w_i$ is its quantized
reconstruction, and the input Hessian $H$ is shared across all output rows.

MACaQ applies a per-channel scale followed by a block-Hadamard rotation. Let
$D=\operatorname{Diag}(t)$ contain the per-channel scales and let $B$ denote a
block-diagonal Hadamard transform with blocks of size $32$. We transform the
activations and weights as
\begin{equation}
X' = XDB,
\qquad
W' = WD^{-1}B.
\end{equation}
Because $B$ is orthogonal, the transformation preserves the unquantized linear
operation:
\begin{equation}
X'{W'}^\top
=
XDB\left(WD^{-1}B\right)^\top
=
XW^\top.
\end{equation}

The corresponding Hessian in the transformed basis is
\begin{equation}
H'={X'}^\top X'
=
B^\top DHD B.
\end{equation}
For the exponent selection and Flip-Refinement steps below, we use a
ridge-regularized version
\begin{equation}
H_f
=
H'
+
\lambda\,
\operatorname{mean}\!\left(\operatorname{diag}(H')\right)I.
\label{eq:ridge-hessian}
\end{equation}
We use $\lambda=1.0$ by default.

At inference time, a MACaQ linear layer computes
\begin{equation}
y
=
\left[(x\odot t)B\right]_{\mathrm{MXFP4}}
\left[\widehat W'^\top\right]_{\mathrm{MXFP4}},
\label{eq:macaq-forward}
\end{equation}
where $\widehat W'$ denotes the MXFP4 reconstruction of $W'$ and
$[\cdot]_{\mathrm{MXFP4}}$ denotes quantization to MXFP4. Thus, all
activation-side transformations remain local to the linear layer.

MACaQ quantizes each layer in four steps: \textbf{(i)} per-channel balancing,
\textbf{(ii)} block-Hadamard rotation, \textbf{(iii)} Hessian-weighted block
exponent selection, and \textbf{(iv)} Flip-Refinement of the FP4 codes.

\subsection{Online Scaling}

The first transform is a per-channel scale $t$. It divides the weight columns
before quantization and multiplies the corresponding activation channels:
\begin{equation}
xW^\top
=
(x\odot t)(W\oslash t)^\top.
\end{equation}
The two operations cancel in the unquantized matrix multiplication, but change
the distributions presented to the weight and activation quantizers.

The aim of this transformation is to reduce the magnitude of outlying
activation channels while simultaneously balancing the weight columns. There
is a natural trade-off: scaling down a given activation channel scales up the
corresponding weight column, and vice versa \cite{smoothquant}. We choose
\begin{equation}
t_j
=
\frac{(\sigma_j^W)^\alpha}
     {(\operatorname{rms}_j^x)^{1-\alpha}},
\qquad
\operatorname{rms}_j^x
=
\sqrt{\frac{H_{jj}}{N}},
\label{eq:online-scale}
\end{equation}
where $\sigma_j^W$ is the standard deviation of weight column $j$ and
$\operatorname{rms}_j^x$ is the RMS magnitude of activation channel $j$,
obtained directly from the diagonal of $H$. We use \hbox{$\alpha=0.75$}, chosen
on a validation set (a subset of wikitext-2 train).

Channels with large activation RMS receive smaller
values of $t_j$ and are therefore attenuated before activation quantization.
Conversely, weight columns with large standard deviation receive larger values
of $t_j$ and are reduced by the division $W\oslash t$. The parameter $\alpha$
controls the trade-off between activation and weight balancing. We clamp the
resulting scales to a finite range for numerical stability.

\subsection{Online Micro-Rotation}

After scaling, we apply a block-Hadamard rotation along the input-channel
dimension. For the block-diagonal orthogonal matrix $B$,
\begin{equation}
xW^\top
=
(xB)(WB)^\top.
\end{equation}
As in prior work \cite{quip}, the Hadamard transform suppresses outliers by
distributing concentrated channel energy across approximately
Gaussian-distributed transformed channels.

We restrict the transform to blocks of size $32$, matching the MXFP4
microscaling block size. This avoids spreading an outlying channel across
multiple independently scaled MXFP4 blocks \cite{duquant++}. The same fixed transform is
applied to the scaled activations online and absorbed into the transformed
weights offline.

Combining scaling and rotation gives
\begin{equation}
x'=(x\odot t)B,
\qquad
W'=(W\oslash t)B.
\end{equation}
MACaQ first applies the per-channel scale and then the
block-Hadamard transform, because the importance weighting of channels must
happen in the unrotated domain.

\subsection{Hessian-Weighted Exponent Selection}
\label{sec:exponent-screen}

In MXFP4, every block shares a power-of-two E8M0 scale. Selecting this scale
presents a trade-off between resolution and clipping: a smaller scale gives
finer resolution around zero but may clip large values, whereas a larger scale
avoids clipping at the cost of coarser quantization.

Before refining the individual FP4 codes, MACaQ screens a small set of candidate
exponents independently for every weight row and MXFP4 block. Let
$\mathcal{B}$ denote one block of transformed weight row $i$. For each candidate
exponent $e$, we round the block to the corresponding E2M1 grid and compute
\begin{equation}
E_{i,\mathcal{B}}(e)
=
\sum_{j\in\mathcal{B}}
(H_f)_{jj}
\left(
W'_{ij}
-
\widehat W'_{ij}(e)
\right)^2.
\label{eq:exponent-screen}
\end{equation}
We select the exponent that minimizes
$E_{i,\mathcal{B}}(e)$.

We evaluate offsets $\{+1,0,-1\}$ around the standard
exponent derived from the block maximum. The
selected block exponent is then held fixed during Flip-Refinement.

\subsection{Flip-Refinement}
\label{sec:flip}

The exponent screen produces an initial round-to-nearest assignment of the FP4
codes. Starting from this assignment, Flip-Refinement minimizes the
ridge-regularized Hessian objective by moving individual codes to adjacent E2M1
levels.

For output row $i$, let
\begin{equation}
r_i=w'_i-\widehat w'_i,
\qquad
g_i=H_f r_i^\top.
\end{equation}
Suppose changing code $(i,j)$ changes its reconstructed value by $\Delta$, such
that
\begin{equation}
\widehat W'_{ij}
\leftarrow
\widehat W'_{ij}+\Delta.
\end{equation}
The residual then changes from $r_i$ to $r_i-\Delta e_j$. The resulting
decrease in the row objective is
\begin{equation}
\operatorname{gain}(i,j,\Delta)
=
2\Delta g_{i,j}
-
\Delta^2(H_f)_{jj}.
\label{eq:flip-gain}
\end{equation}
A positive gain implies an improving move. After accepting the
flip, the gradient is updated incrementally:
\begin{equation}
g_i
\leftarrow
g_i-\Delta(H_f)_{:,j}.
\label{eq:flip-update}
\end{equation}
This update avoids recomputing the complete matrix-vector product after each
flip.

The E2M1 grid is non-uniform, so $\Delta$ depends on the current code, the
chosen direction, and the fixed block scale. At every iteration, we evaluate
the adjacent levels of every code and select the single highest-gain move for
each output row. The output rows are independent and are processed in parallel.
Rows for which no positive-gain move remains are removed from the active set and
require no further work. Algorithm~\ref{alg:flip-refine} summarizes the
procedure.

\begin{algorithm}[tb]
\caption{Flip-Refinement for one linear layer}
\label{alg:flip-refine}
\textbf{Require}: transformed weights $W'\in\mathbb{R}^{m\times n}$;
regularized Hessian $H_f$; initial FP4 codes $Q$; fixed block scales $s$;
maximum iterations $K$\\
\textbf{Ensure}: refined FP4 codes $Q$
\begin{algorithmic}[1]
\FOR{$i\leftarrow 1$ \textbf{to} $m$ \textbf{ in parallel}}
    \STATE $\widehat w'_i\leftarrow dq(Q_i;s_i)$
    \STATE $r_i\leftarrow w'_i-\widehat w'_i$
    \STATE $g_i\leftarrow H_f r_i^\top$
\ENDFOR
\STATE $\mathcal{A}\leftarrow\{1,\ldots,m\}$
\FOR{$k\leftarrow 1$ \textbf{to} $K$}
    \FOR{$i\in\mathcal{A}$ \textbf{ in parallel}}
        \STATE $(j_i,\Delta_i)\leftarrow
        \arg\max_{j,\Delta}%
        \left[2\Delta g_{i,j}-\Delta^2(H_f)_{jj}\right]$
        \STATE $\gamma_i\leftarrow
        2\Delta_i g_{i,j_i}-\Delta_i^2(H_f)_{j_i j_i}$
    \ENDFOR
    \STATE $\mathcal{I}\leftarrow
    \{i\in\mathcal{A}:\gamma_i>0\}$
    \IF{$\mathcal{I}=\emptyset$}
        \STATE \textbf{break}
    \ENDIF
    \FOR{$i\in\mathcal{I}$ \textbf{ in parallel}}
        \STATE $Q_{i,j_i}\leftarrow$ selected adjacent code
        \STATE $g_i\leftarrow g_i-\Delta_i(H_f)_{:,j_i}$
    \ENDFOR
    \STATE $\mathcal{A}\leftarrow\mathcal{I}$
\ENDFOR
\STATE \textbf{return} $Q$
\end{algorithmic}
\end{algorithm}

Here, $\Delta \in \mathcal{N}(Q_{ij};s_{ij})$ are the reconstructed-value changes
associated with moving one level upward or downward on the local FP4 grid (where possible).

Each iteration accepts at most one flip per output row, so $K$ is also the
maximum number of accepted flips per row. In practice, many rows converge
before reaching this budget. 

\subsection{Cost of Online Transforms}
\label{sec:overhead-theory}

The activation-side pipeline of a MACaQ linear layer is:
\textbf{(i)} apply the per-channel scale,
\textbf{(ii)} apply the block-Hadamard transform, and
\textbf{(iii)} quantize the result to MXFP4. Activation quantization is required
by any W4A4 MXFP4 implementation, so the additional operations introduced by
MACaQ are only the scale and the micro-rotation.

For a layer processing $T$ tokens with input dimension $n$ and output dimension
$m$, a block-Hadamard transform of size $B$ requires approximately
$Tn\log_2 B$ additions, while the scale requires $Tn$ multiplications. By
comparison, the matrix multiplication performs approximately $2Tmn$
arithmetic operations. The ratio is therefore
\begin{equation}
\frac{C_{\mathrm{online}}}{C_{\mathrm{matmul}}}
=
\frac{Tn(\log_2 B+1)}{2Tmn}
=
\frac{\log_2 B+1}{2m}.
\end{equation}
For $B=32$, this becomes
\begin{equation}
\frac{C_{\mathrm{online}}}{C_{\mathrm{matmul}}}
=
\frac{3}{m}.
\end{equation}

Both online transforms are applied once to each input and reused across all
$m$ output channels. Their cost is therefore independent of $m$, whereas the
cost of the matrix multiplication grows linearly with $m$. For example, the
smallest output dimension among the quantized linear layers of Llama-3-8B is
$m=1024$, giving an arithmetic-operation ratio below $0.3\%$.

This estimate describes arithmetic work rather than wall-clock performance.
The measured overhead also depends on kernel fusion, memory traffic, launch
overhead, and the relative throughput of FP16 and MXFP4 operations. We therefore
measure the actual kernel-level and end-to-end overhead in
later sections. Consistent with the
analysis above, the relative overhead decreases with layer and model size.

\subsection{Quantization Procedure}

MACaQ preserves the layer-wise modularity of the underlying network: each
linear layer has its own scaling factors, block rotation, Hessian, block
exponents, and FP4 codes. Quantizing one linear layer never requires modifying a
neighboring normalization layer, projection, or residual branch.

Forward hooks accumulate
\begin{equation}
H_\ell
=
\sum_k X_{\ell,k}^\top X_{\ell,k}
\end{equation}
for every linear layer $\ell$ over chunks of the calibration sequence. After
collecting these statistics, each linear layer is independently scaled,
rotated, exponent-screened, and refined. The original layer is then replaced by
its packed MXFP4 implementation.

Only the layer currently being processed needs to reside on the
GPU (and for the calibration, outputs of parallel layers). 
The remaining model weights remain in host memory. This streaming
procedure reduces peak device memory and allows the same
quantization implementation to be reused across model architectures; all calibration
and quantization logic is attached locally to linear layers.

\subsection{Implementation Details}

MACaQ offers two levers for trading quantization time against output
quality: the number of calibration tokens and the maximum
number $K$ of Flip-Refinement iterations. %

We produce the Pareto curve in Fig.~\ref{fig:pareto} by sweeping these two
parameters. Quality begins to plateau at approximately $2\times 10^6$
calibration tokens and $K=512$ Flip-Refinement iterations. For fair comparison we use $K=128$ and $2\times10^5$ calibration tokens for MACaQ in the experimental section (ca. 150s to quantize Llama3-8B).

Besides this, our default configuration uses block size $32$, exponent offsets
$\{+1,0,-1\}$, scale-balancing parameter $\alpha=0.75$, and ridge coefficient
$\lambda=1.0$. The parameters $\alpha$ and $\lambda$ were chosen with 
a wikitext train set validation
run on Llama3-8B (sensitivity plot in appendix). The Flip-Refinement implementation parallelizes across output
rows and removes converged rows from the active set. We additionally compile
the repeated refinement step with \texttt{torch.compile}; the initial
compilation cost is amortized across layers with matching shapes.

\section{Experiments}

To show the effectiveness of MACaQ, we consider several different metrics. For
comparison with literature baselines, we use model perplexity, as it is widely
reported in quantization work and provides a continuous and comparably reliable measure of model
quality. 

We further report results on generative benchmarks. In generative
settings, quantization errors may accumulate over generated sequences in ways
that are not visible in fixed-text metrics such as perplexity, and these tasks
are closer to many practical use cases.
To illustrate the ease of generalization to new architectures, we also apply
MACaQ to an image diffusion model. Finally, we measure the throughput of MACaQ
relative to standard FP16 and other common baselines, both at the kernel level
and end-to-end, and perform an ablation over the main components of MACaQ.

The models we evaluate are drawn from the Llama-3 \cite{grattafiori2024llama}, Qwen3 \cite{yang2025qwen3}, and
Qwen3.5 \cite{team2026qwen3} families. We evaluate perplexity on Llama models for
comparability with prior work and use the reasoning-capable Qwen models for
generative reasoning evaluations. For image diffusion, we use FLUX.2 Klein 4B
\cite{flux-2-2025}, as it is openly available.

Note that for MoE models we only quantize expert layers (also for baselines) 
otherwise we quantize all weights and activations, 
but not the embeddings and lm-head (as is customary). 

All experiments are run on a recent GPU with 504 TFLOPS (FP16), 96 GB memory and 1.79 TB/s memory bandwidth and use OCP MXFP4 format.

\begin{figure}
\centering
\includegraphics[width=0.85\linewidth]{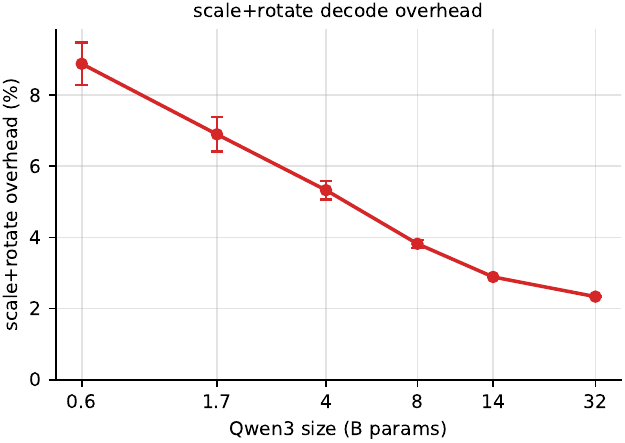}
\caption{Decode overhead introduced by online rotation and scaling (batch 1, context 512). The
overhead decreases with model size and reaches approximately $2.3\%$ for
Qwen3-32B. Error bars show the standard error.}
\label{fig:decode_overhead_sr}
\end{figure}

\subsection{Perplexity}

For perplexity measurements, we follow related work and calibrate on the
WikiText-2 training set. We evaluate on the WikiText-2 \cite{merity2016pointer_wikitext} and C4 \cite{raffel2020exploring_c4} test sets.
WikiText-2 can therefore be understood as an in-distribution evaluation with
respect to the calibration data, while C4 provides an out-of-distribution
evaluation. We use a context length of $2048$. The results in Tab.~\ref{tab:ppl} and Fig.~\ref{fig:pareto}
show that MACaQ matches or outperforms the baselines also in terms of
quality, not only speed. 

\begin{table*}
\centering
\caption{OCP MXFP4 W4A4, WikiText2 / C4 perplexity ($\downarrow$). Best quantized result per column in bold, second best
underlined.
}
\label{tab:ppl}
\small
\setlength{\tabcolsep}{4.5pt}
\resizebox{\linewidth}{!}{%
\begin{tabular}{l rr rr rr rr rr rr}
\toprule

 & \multicolumn{2}{c}{Llama-3-8B} & \multicolumn{2}{c}{Llama-3.2-3B}
 & \multicolumn{2}{c}{Llama-3-8B-Inst.} & \multicolumn{2}{c}{Llama-3.1-8B-Inst.} & \multicolumn{2}{c}{Qwen3-32B} & \multicolumn{2}{c}{Qwen3.5-35B-A3B}\\
\cmidrule(lr){2-3}\cmidrule(lr){4-5}\cmidrule(lr){6-7}\cmidrule(lr){8-9}\cmidrule(lr){10-11}\cmidrule(lr){12-13}
Method & Wiki2 & C4 & Wiki2 & C4 & Wiki2 & C4 & Wiki2 & C4 & Wiki2 & C4 & Wiki2 & C4\\
\midrule
FP16   & 6.14 & 9.46  & 7.81  & 11.34 & 8.31  & 13.03 & 7.16 & 11.41 & 7.62 & 12.45 & 6.55 & 10.98\\
\midrule
QuaRot (RTN)      & 9.34 & 15.53 & 12.40 & 18.63 & 11.74 & 19.21 & 10.00 & 16.48 & 10.19 & 16.59 & 6.77 & 11.38\\
QuaRot (GPTQ)     & 7.65 & 12.62 & 9.65 & 15.11 & 9.62 & 15.85 & 8.58 & 14.49 & 8.46 & 13.42& 6.79 & 11.33\\
FlatQuant         & 7.21 & 11.65 & 9.00  & 14.87 & 9.25  & 15.27 & \underline{8.10} & 13.40 & 8.11 & 13.28 & 6.87 & 11.57 \\
MR-GPTQ           & \underline{7.13} & \underline{11.37} & \underline{8.92}  & \underline{13.51} & 9.13  & \underline{14.62} & 8.14 & \underline{13.24} & 8.15 & 13.27 & 6.75 & \underline{11.22} \\
DuQuant++         & 7.48 & 11.84 & 9.36  & 14.19 & 9.37  & 14.86 & 8.41 & 13.67 & 8.27 & 13.33 & \underline{6.73} & 11.28\\
DuQuant++ (GPTQ)  & 7.17 & 11.66 & 9.02  & 13.92 & \underline{9.04}  & 14.64 & 8.18 & 13.48 & \underline{8.07} & \underline{13.15} & 6.75 & 11.26 \\
\midrule
\textbf{MACaQ (ours)} & \textbf{6.90} & \textbf{11.01} & \textbf{8.61} & \textbf{13.07} & \textbf{8.90} & \textbf{14.41} & \textbf{7.88} & \textbf{12.91} & \textbf{7.94} & \textbf{12.93} & \textbf{6.66} & \textbf{11.21} \\
\midrule
\bottomrule
\end{tabular}}
\end{table*}

\subsection{Large-Model Quantization}

MACaQ can be used to efficiently quantize models that are too large to reside
on the target GPU at full precision. To demonstrate this, we quantize
Qwen3.5-122B-A10B on a machine with a single $96$GB GPU. The model contains
$122$B total parameters, and its FP16
weights require ca. $244$GB.

MACaQ processes the model with lazy loading of layers and reaches a
GPU-memory usage of $32.5$GB. Quantization
takes $64$ minutes (with 32 flips and 512K calibration tokens) 
and produces a WikiText-2 perplexity of $5.14$,
compared with $5.06$ for the full-precision model. The resulting
MXFP4 model requires $74$GB of GPU memory and can run on the same
$96$GB GPU. Thus, MACaQ can quantize a model that does not fit on the target GPU
in its original precision into one that does.

\subsection{Generative Language Tasks}

Generative benchmarks are closer to many practical workloads and
have the potential to reveal accumulation effects that may only appear when a
model generates a sequence of tokens. They also avoid the limitations
of multiple-choice QA evaluation highlighted by \citet{notaccuracy}.

We evaluate MACaQ-quantized dense Qwen3 and Qwen3.5 MoE models on IFEval \cite{zhou2023instructionfollowingevaluationlargelanguage_ifeval} and
MATH-500 \cite{hendrycksmath2021_math500}. MATH-500 contains free-form, non-multiple-choice mathematics
problems that require reasoning chains to answer correctly. IFEval tests the instruction
following capabilities of models in generation. 

Tab.~\ref{tab:reasoning} shows that MACaQ performs well on these tasks.

\begin{table*}
\centering
\caption{OCP MXFP4 W4A4. \textbf{IFEval} strict prompt-level accuracy
($\uparrow$, average over $k{=}10$; {\tiny$\pm$} across-seed standard error).
\textbf{MATH500} accuracy ($\uparrow$, average over $k{=}10$; {\tiny$\pm$} across-seed standard error).
Best quantized result per column in bold,
second best underlined; also all results within one standard error of the baseline are bolded. }
\label{tab:reasoning}
\small
\setlength{\tabcolsep}{4.5pt}
\begin{tabular}{l cc cc}
\toprule
 & \multicolumn{2}{c}{Qwen3-32B} & \multicolumn{2}{c}{Qwen3.5-35B-A3B} \\
\cmidrule(lr){2-3}\cmidrule(lr){4-5}
Method & IFEval & MATH500 & IFEval & MATH500 \\
\midrule
Original (FP16)   & 83.60\%{\tiny$\pm$0.25} &  97.78\%{\tiny$\pm$0.10} & 85.29\%{\tiny$\pm$0.25} & 99.38\%{\tiny$\pm$0.06} \\
\midrule
QuaRot (RTN)      &  78.46\%{\tiny$\pm$0.23} & 93.65\%{\tiny$\pm$0.20}  & \textbf{85.41}\%{\tiny$\pm$0.33}  & 98.78\%{\tiny$\pm$0.10} \\
QuaRot (GPTQ)     &  82.35\%{\tiny$\pm$0.29} & 96.98\%{\tiny$\pm$0.19}  &  \textbf{85.60\%}{\tiny$\pm$0.39} & 98.60\%{\tiny$\pm$0.10} \\
MR-GPTQ           & \underline{82.81\%}{\tiny$\pm$0.27} & \textbf{97.44}\%{\tiny$\pm$0.16}  &  \textbf{85.43\%}{\tiny$\pm$0.24} & \underline{98.86\%}{\tiny$\pm$0.07} \\
DuQuant++ (RTN)        &  81.57\%{\tiny$\pm$0.29}  & 96.82\%{\tiny$\pm$0.21} & 85.29\%{\tiny$\pm$0.20} & 98.58\%{\tiny$\pm$0.11}\\
DuQuant++ (GPTQ)  &  \underline{82.81\%}{\tiny$\pm$0.27}  & 96.90\%{\tiny$\pm$0.14}  & \textbf{85.59\%}{\tiny$\pm$0.21} & 98.50\%{\tiny$\pm$0.11} \\
\midrule
\textbf{MACaQ (ours)} & \textbf{83.20\%}{\tiny$\pm$0.26} & \underline{97.26}\%{\tiny$\pm$0.11}  &  \textbf{85.42}\%{\tiny$\pm$0.40}  & \textbf{98.90\%}{\tiny$\pm$0.09} \\
\midrule
\bottomrule
\end{tabular}
\end{table*}

\subsection{Image Diffusion}

One of the key distinguishing features of MACaQ is that it can be applied to
new architectures with minimal effort because it treats each linear layer in
isolation. We demonstrate this by quantizing the linear layers of FLUX.2 Klein
4B \cite{flux-2-2025} to MXFP4. 

This model also offers an official  NVFP4 variant\footnote{for a fair comparison we quantize the same layers that this official NVFP4 model quantizes with MACaQ}, 
which provides a strong baseline. NVFP4
uses microscaling groups of $16$, rather than the $32$ used by MXFP4, and
requires approximately $4.5$ rather than $4.25$ bits per element. 

\begin{figure}
    \centering
    \includegraphics[width=0.9\linewidth]{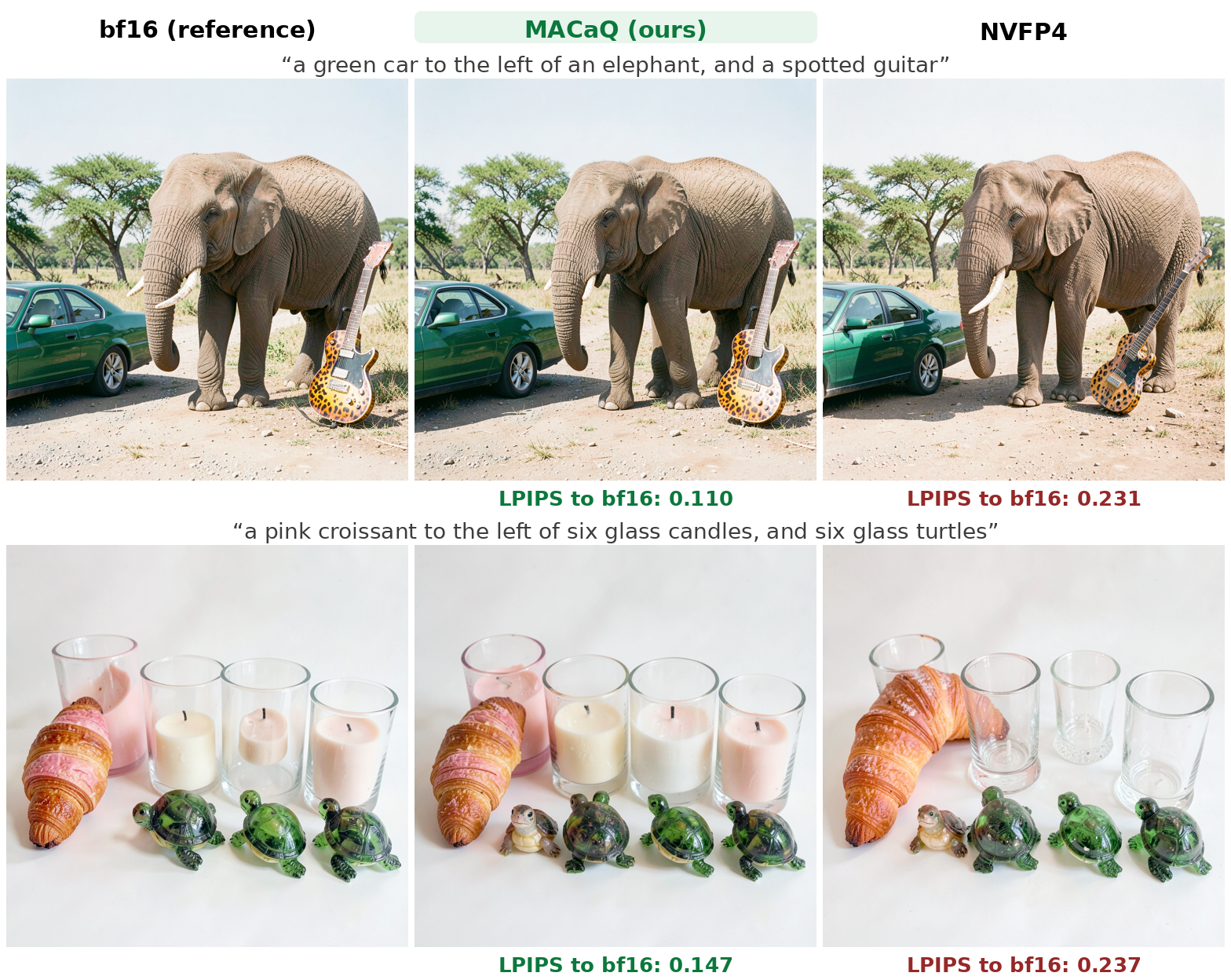}
    \caption{Comparison of FLUX2-klein-4B under MACaQ MXFP4 vs. the official 
    NVFP4 quantization. MACaQ is more similar to the high precision model; note e.g. the 
    shadow's orientation in the first image and the empty glasses in the second.
     Quantitative results in Tab.~\ref{tab:flux}.}
    \label{fig:flux}
\end{figure}

We find that MACaQ produces a more faithful quantized model with
respect to the original BF16 model when prompted with GenEval2 prompts \cite{kamath2025geneval}. 
In Table~\ref{tab:flux}, %
we report pairwise
metrics ranging from signal-level similarity (PSNR), through perceptual
similarity (LPIPS), to semantic similarity (DINOv3). MACaQ is closer to the
original BF16 model across all these metrics. For a qualitative illustration see
Fig.~\ref{fig:flux}.

\begin{table*}[t]
\small
\centering
\caption{\textbf{FLUX.2-klein-4B}: MACaQ vs.\ the official NVFP4 QAD
checkpoint. GenEval2 (800 prompts), $1024^2$, 6 steps, DiT-only W4A4.
PSNR/SSIM/LPIPS and DINOv3 (ViT-L/16) feature cosine are paired against the
BF16 output (same prompt and seed). Peak VRAM at batch~1.
MACaQ is closer to BF16 on all four fidelity metrics and ties on VRAM.}
\label{tab:flux}
\setlength{\tabcolsep}{5pt}
\begin{tabular}{llccccccc}
\toprule
Method & Format & bits/w & PSNR\,$\uparrow$ & SSIM\,$\uparrow$ & LPIPS\,$\downarrow$ & DINO$_{\text{global}}$\,$\uparrow$ & DINO$_{\text{patch}}$\,$\uparrow$ & VRAM\,(GB)\,$\downarrow$ \\
\midrule
Full Precision      & BF16       & 16.0 & - & - & - & - & - & 18.5 \\
\textbf{MACaQ (ours) }& MXFP4  & 4.25 & \textbf{15.04} & \textbf{0.697} & \textbf{0.218} & \textbf{0.882} & \textbf{0.947} & \textbf{13.1} \\
Official NVFP4    & NVFP4  & 4.50 & 14.85 & 0.676 & 0.235 & 0.878 & 0.937 & 13.3 \\
\midrule
\bottomrule
\end{tabular}
\end{table*}

\subsection{Kernel-Level Speed}
\label{sec:kernel-speed}

In Table~\ref{tab:kernel-speed}, we compare the speed of a single linear layer
across different batch sizes and matrix dimensions. We compute
$\mathbf{X}\mathbf{W}^\top$ using square weight matrices
$\mathbf{W}\in\mathbb{R}^{N\times N}$, where
$N\in\{4096,16384\}$, and inputs
$\mathbf{X}\in\mathbb{R}^{B\times L\times N}$, where the batch size
$B\in\{1,16\}$ and context length $L\in\{512,16384\}$. We compare our MACaQ MXFP4 linear layer with, first, the same kernel with online
scaling and rotation disabled; second, the Marlin W4A16 kernel \cite{frantar2025marlin}; and,
finally, a FP16 baseline.

\begin{table}[t]
\centering
\caption{Kernel-level speed of one linear layer $\mathbf{X}\cdot\mathbf{W}^\top$, $\mathbf{W}\in\mathbb{R}^{N\times N}$, $\mathbf{X}\in\mathbb{R}^{B\times L\times N}$ ($M=B\cdot L$ rows; $L=1$ is decode). Median as \textbf{speedup relative to fp16} ($\uparrow$; fp16$\equiv1.00$). Standard Error is omitted as it is below 0.02 in all cases. MACaQ is the MXFP4 W4A4 kernel with online per-channel scaling and block-Hadamard rotation; $-$SR disables both; Marlin is the W4A16 kernel of \citet{frantar2025marlin}. ``SR cost'' is the relative slowdown from enabling scale+rotation.}%
\label{tab:kernel-speed}
\small
\setlength{\tabcolsep}{5pt}
\resizebox{\linewidth}{!}{%
\begin{tabular}{ccc cccc c}
\toprule
$N$ & $B$ & $L$ & MACaQ & $-$SR & Marlin & FP16 & SR cost \\
\midrule
\multirow{4}{*}{\rotatebox{90}{4096}} & 1 & 1 & $\textbf{1.31}{\scriptstyle\,}$ & $1.31{\scriptstyle\,}$ & $\textbf{1.31}{\scriptstyle\,}$ & $1.00$ & <$1\%$ \\
 & 1 & 16384 & $\textbf{1.75}{\scriptstyle\,}$ & $1.86{\scriptstyle\,}$ & $1.02{\scriptstyle\,}$ & $1.00$ & 6.4\% \\
 & 16 & 1 & $\textbf{1.75}{\scriptstyle\,}$ & $1.75{\scriptstyle\,}$ & $\textbf{1.75}{\scriptstyle\,}$ & $1.00$ & <$ 1\%$ \\
 & 16 & 16384 & $\textbf{1.50}{\scriptstyle\,}$ & $1.57{\scriptstyle\,}$ & $1.01{\scriptstyle\,}$ & $1.00$ & 4.1\% \\
\midrule
\multirow{4}{*}{\rotatebox{90}{16384}} & 1 & 1 & $5.04{\scriptstyle\,}$ & $5.11{\scriptstyle\,}$ & $\textbf{8.26}{\scriptstyle\,}$ & $1.00$ & 1.3\% \\
 & 1 & 16384 & $\textbf{2.01}{\scriptstyle\,}$ & $2.05{\scriptstyle\,}$ & $1.00{\scriptstyle\,}$ & $1.00$ & 1.6\% \\
 & 16 & 1 & $4.47{\scriptstyle\,}$ & $4.55{\scriptstyle\,}$ & $\textbf{6.19}{\scriptstyle\,}$ & $1.00$ & 1.7\% \\
 & 16 & 16384 & $\textbf{1.83}{\scriptstyle\,}$ & $1.85{\scriptstyle\,}$ & $0.98{\scriptstyle\,}$ & $1.00$ & 0.9\% \\
 \midrule
\bottomrule
\end{tabular}}
\end{table}

The MACaQ kernel reaches a substantial speedup over FP16, while
the overhead of online scaling and rotation remains low. For the smaller layer,
the overhead is at most approximately $6\%$, and for the larger layer it
remains below $3\%$. As expected from the theoretical analysis, the
relative overhead decreases for larger layers.

\subsection{End-to-End Speed}
\label{sec:e2e-speed}

On Qwen3-32B at batch 1, our MACaQ MXFP4 implementation decodes at 69.8 tok/s against 22.8 for FP16 (3.06x); at batch 256 the gain is 5252.8 vs 3061.5 tok/s. Per-batch-size numbers for both models are in the appendix. MACaQ achieves up to $3\times$ end-to-end speed-up over FP16.
Generally, speed-ups are comparable to Marlin \cite{frantar2025marlin} 
in the memory-bound case and outperform Marlin in compute-bound settings. 

In Fig.~\ref{fig:decode_overhead_sr}, we isolate the overhead caused by online
rotation and scaling. For the small Qwen3-0.6B model, this overhead is
approximately $8\%$(i.e., MACaQ still provides a substantial speedup over FP16). 
As explained in the Method section, the overhead
decreases with model size, reaching approximately $2\%$ for Qwen3-32B.

\subsection{Ablation}

In Table~\ref{tab:ablation}, we show the contributions of the main components
of MACaQ. Each component provides an improvement of a similar order of
magnitude, and their combination achieves the lowest quantization error.

\begin{table}[t]
\centering
\caption{Additive contribution of each MACaQ component on Llama-3-8B (MXFP4 W4A4), WikiText-2 perplexity at 512 Flips and 200'000 calibration tokens.}
\label{tab:ablation}
\begin{tabular}{lcc}
\toprule
Configuration & Wiki2 $\downarrow$ & $\Delta$ \\
\midrule
Full Precision (FP16)       & 6.14 & --      \\
RTN                         & 8.28 & --      \\
+ Hessian exponent search   & 8.09 & $-0.19$ \\
+ micro-rotation            & 7.65 & $-0.44$ \\
+ online second scale       & 7.08 & $-0.57$ \\
+ flip refinement (= MACaQ) & 6.87 & $-0.21$ \\
\midrule
\bottomrule
\end{tabular}
\end{table}

\section{Related Work}

\paragraph{Hessian-Based Weight Compression}
Much of the current \emph{second-order} quantization literature builds on
Optimal Brain Damage (OBD) \cite{lecun1989optimal} and Optimal Brain
Surgeon (OBS) \cite{OBS} that introduced Hessian-based model compression. 
Optimal Brain Compression (OBC) \cite{frantar2022optimal} adapted this framework to
layer-wise compression and introduced Optimal Brain Quantization (OBQ), using
the layer-input Hessian $H=X^\top X$ to minimize output reconstruction error.
GPTQ \cite{GPTQ} subsequently made this approach scalable to large language
models by fixing the quantization order, allowing the same Hessian factorization
to be reused across output rows. These methods assign weights sequentially and
compensate the remaining unquantized weights. In contrast, MACaQ begins from a
cheap RTN assignment and then revisits the individual
codes through incremental Hessian-aware updates.

\paragraph{Coordinate-Descent Quantization}
QuantEase \cite{behdin2023quanteaseoptimizationbasedquantizationlanguage} and CDQuant 
\cite{nair2024cdquantgreedycoordinatedescent} also formulate
quantization (pure weight-quantization) as coordinate descent and improve over one-shot assignments such
as GPTQ by revisiting quantized weights. 
Our Flip-Refinement is designed primarily
for speed in weight and activation quantization. 
It keeps each MXFP4 block exponent fixed and considers
only the two E2M1 levels adjacent to the current code, allowing closed-form evaluation of the gain of each move.

\paragraph{Basis Transforms}
Basis transforms have become a central tool for reducing weight and activation
quantization error. QuIP \cite{quip} popularized incoherence processing through
Hadamard transforms for weight quantization, and QuaRot \cite{quarotb} extended
Hadamard rotations to weight--activation quantization. SmoothQuant
\cite{smoothquant} used element-wise scaling to balance errors between 
activations and weights. Later work has optimized the
rotation itself \cite{spinquant, duquant}, general per-layer affine maps, as in FlatQuant \cite{flatquant}, or an element-wise scale, including weight-only quantization with AWQ \cite{AWQ} and
calibration-free weight quantization with SINQ \cite{SINQ}

For microscaling formats, block-local rotations are often preferable to
full-channel rotations, e.g. MR-GPTQ \cite{egiazarian2026bridging_MR-GPTQ} and \cite{shao2025block_brq, duquant++}, as they avoid spreading an
outlier across multiple independently scaled quantization blocks. Other work
has studied how these transforms interact: \cite{federici2026dissecting} analyze the
sources of error in weight--activation quantization and recommend combining
rotation and scaling, while \cite{sanjeet2026pushing} show that reducing
concentrated activation mass makes Hadamard transforms more effective
preconditioners.

\section{Limitations}
The core design decision of MACaQ, keeping everything layer-local, means that cross-layer error compensation is impossible. A secondary pass with cross-layer statistics could be added on top, but would likely incur the slower quantization times of other methods. 

In our MoE experiments we restricted ourselves to expert quantization only due to practicality considerations. 

For the smallest model tested (0.6B parameters) the online basis transform overhead climbs to $8\%$, which starts to be relevant for some settings.

\section{Conclusion}

We have introduced MACaQ, a fast and easy to apply MXFP4 quantization 
method for models based on linear layers. MACaQ matches or outperforms
state of the art methods in quality and is faster. The key concepts
that enable this are a light-weight Flip-Refinement to replace GPTQ, 
a novel Hessian-aware shared-scale optimization, and light-weight
\emph{online} basis transforms that enable the quantization processing to be 
local to each linear layer. All these features combined enable us to
quantize a 122B Qwen3.5 model in 64 minutes while using only 32.5GB 
GPU RAM.

\clearpage
\appendix

\section{Appendix}
\subsection{End to end speed table}
Here we show detailed end to end speed results in Tab.~\ref{tab:e2e-speed}.

\begin{table}[h]
\centering
\caption{End-to-end throughput (median tok/s; 512-token prompt, vLLM graph mode) across batch sizes. Prefill$=$prompt$/$TTFT, decode$=$per-token rate, aggregated over the batch. MACaQ$-$SR disables the online scale$+$rotation (speed-only ablation).}
\label{tab:e2e-speed}
\small
\setlength{\tabcolsep}{6pt}
\begin{tabular}{l cc cc}
\toprule
 & \multicolumn{2}{c}{Llama-3-8B} & \multicolumn{2}{c}{Qwen3-32B} \\
\cmidrule(lr){2-3}\cmidrule(lr){4-5}
Method & Prefill & Decode & Prefill & Decode \\
\midrule
\multicolumn{5}{l}{\textit{Batch 1}} \\
bf16 & 38{,}171 & 93.2 & 11{,}027 & 22.8 \\
Marlin W4A16 & \textbf{71{,}760} & 223.9 & \textbf{29{,}307} & \textbf{71.8} \\
\textbf{MACaQ (ours)} & 70{,}547 & \textbf{227.9} & 28{,}473 & 69.8 \\
\cmidrule(lr){1-5}
\textit{MACaQ $-$SR} & 72{,}852 & 232.9 & 28{,}364 & 69.4 \\
\midrule
\multicolumn{5}{l}{\textit{Batch 16}} \\
bf16 & 236{,}101 & 1383.9 & 57{,}621 & 342.7 \\
Marlin W4A16 & \textbf{442{,}456} & \textbf{3034.9} & \textbf{147{,}688} & \textbf{994.6} \\
\textbf{MACaQ (ours)} & 436{,}001 & 2842.1 & 143{,}225 & 949.4 \\
\cmidrule(lr){1-5}
\textit{MACaQ $-$SR} & 467{,}704 & 3058.2 & 147{,}370 & 976.9 \\
\midrule
\multicolumn{5}{l}{\textit{Batch 32}} \\
bf16 & 393{,}524 & 2471.6 & 107{,}240 & 652.1 \\
Marlin W4A16 & 599{,}408 & \textbf{5251.5} & 203{,}672 & \textbf{1792.6} \\
\textbf{MACaQ (ours)} & \textbf{650{,}692} & 5028.2 & \textbf{236{,}880} & 1756.7 \\
\cmidrule(lr){1-5}
\textit{MACaQ $-$SR} & 665{,}454 & 5426.4 & 243{,}237 & 1804.2 \\
\midrule
\multicolumn{5}{l}{\textit{Batch 64}} \\
bf16 & 607{,}508 & 4527.5 & 171{,}756 & 1234.7 \\
Marlin W4A16 & 748{,}422 & 7750.3 & 226{,}696 & 2697.8 \\
\textbf{MACaQ (ours)} & \textbf{904{,}177} & \textbf{7840.2} & \textbf{294{,}347} & \textbf{2709.8} \\
\cmidrule(lr){1-5}
\textit{MACaQ $-$SR} & 954{,}273 & 8418.5 & 305{,}896 & 2836.7 \\
\midrule
\multicolumn{5}{l}{\textit{Batch 128}} \\
bf16 & 816{,}722 & 7566.8 & 224{,}958 & 2086.1 \\
Marlin W4A16 & 915{,}282 & 9502.0 & 266{,}406 & 3170.0 \\
\textbf{MACaQ (ours)} & \textbf{1{,}087{,}913} & \textbf{11158.3} & \textbf{323{,}884} & \textbf{3983.3} \\
\cmidrule(lr){1-5}
\textit{MACaQ $-$SR} & 1{,}112{,}510 & 11345.3 & 333{,}024 & 4080.4 \\
\midrule
\multicolumn{5}{l}{\textit{Batch 256}} \\
bf16 & 965{,}564 & 10146.8 & 243{,}291 & 3061.5 \\
Marlin W4A16 & 1{,}020{,}529 & 11220.3 & 276{,}373 & 3573.3 \\
\textbf{MACaQ (ours)} & \textbf{1{,}157{,}387} & \textbf{14207.6} & \textbf{373{,}713} & \textbf{5252.8} \\
\cmidrule(lr){1-5}
\textit{MACaQ $-$SR} & 1{,}143{,}268 & 14899.2 & 375{,}743 & 5402.9 \\
\midrule
\bottomrule
\end{tabular}
\end{table}

\section{Pareto-Plot Numerical Details}
Here we report the numbers relating to the plot in Fig.~\ref{fig:pareto} in 
Tab.~\ref{tab:pareto-quant-time}.

\begin{table}[t]
\centering
\caption{Quantization cost vs.\ quality on Llama-3-8B, MXFP4 W4A4.
\emph{Calib. Blocks} is the calibration budget in units of 2048 tokens: for the
GPTQ-based baselines one block is exactly one $2048$-token calibration
sequence, while MACaQ consumes a flat token stream (its rungs are round decimal
token counts, hence the fractional blocks). PPL is WikiText-2 perplexity; FP16
is 6.14. The $^*$ configuration uses only rotation and exponent screen.}
\label{tab:pareto-quant-time}
\setlength{\tabcolsep}{4pt}
\begin{tabular}{@{}llrrr@{}}
\toprule
Method & Flips & Calib. Blocks & Time (s) & PPL \\
\midrule
FP16 & --  & --  & --    & 6.14 \\
RTN  & --  & 0   & 2.2   & 8.29 \\
\midrule
\multirow{10}{*}{\shortstack[l]{MACaQ\\(ours)}}
 & 0$^{*}$ & 0     & 5.5   & 7.64 \\
 & 0       & 1     & 14.8  & 7.11 \\
 & 16      & 2     & 24.3  & 7.01 \\
 & 32      & 4     & 31.3  & 6.97 \\
 & 64      & 7.8   & 50.8  & 6.96 \\
 & 64      & 16    & 61.8  & 6.93 \\
 & 128     & 31.3  & 88.1  & 6.91 \\
 & 256     & 125   & 193.1 & 6.87 \\
 & 512     & 488.3 & 431.0 & 6.86 \\
 & 512     & 976.6 & 636.9 & 6.85 \\
\midrule
\multirow{5}{*}{MR-GPTQ}
 & -- & 4   & 219.5 & 7.39 \\
 & -- & 16  & 230.3 & 7.18 \\
 & -- & 32  & 242.5 & 7.15 \\
 & -- & 64  & 267.9 & 7.11 \\
 & -- & 128 & 319.1 & 7.11 \\
\midrule
\multirow{5}{*}{\shortstack[l]{DuQuant++\\(GPTQ)}}
 & -- & 4   & 308.5 & 7.57 \\
 & -- & 16  & 347.4 & 7.27 \\
 & -- & 32  & 381.3 & 7.19 \\
 & -- & 64  & 457.5 & 7.19 \\
 & -- & 128 & 624.4 & 7.16 \\
 \midrule
\bottomrule
\end{tabular}
\end{table}

\subsection{Hyperparameter sensitivity}
\begin{figure*}
    \centering
    \includegraphics[width=.8\linewidth]{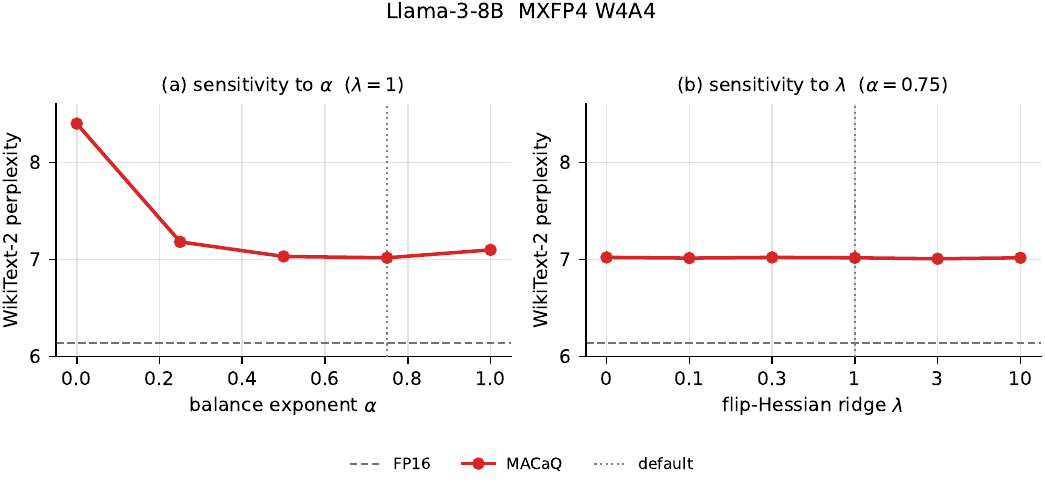}
    \caption{Hyperparameter sensitivity of MACaQ on Llama3-8B in the 4K token 16 flips configuration. Because
    the $\lambda$ curve is flat, we choose a conservative value of $1.0$ to prevent possible under-calibration issues.}
    \label{fig:hparams}
\end{figure*}

We find that MACaQ is not very sensitive to hyperparameter selection (see Fig.~\ref{fig:hparams})

\subsection{OCP MXFP4 compliance}
The OCP MXFP4 specification requires the shared block scale to be a power
of two (E8M0). The DuQuant++ codebase provides both
\path{cast_to_eBm0()} and \path{cast_to_eBm0_improved()}, the
latter permitting fractional-power-of-two scales, which are more expressive
but not representable in the E8M0 scale field. We use a compliant cast
throughout, for all methods including our own, which accounts for a small
discrepancy with some of the numbers reported in the DuQuant++ paper.

\subsection{Additional Generated Images}
Here we add further images generated by the FLUX2 baseline and quantized models for GenEval2\footnote{from \url{https://github.com/facebookresearch/GenEval2} under CC BY-NC 4.0 license} prompts and some simple photographic prompts, chosen to highlight the advantages of our model. 

\begin{figure*}
    \centering
    \includegraphics[width=\linewidth]{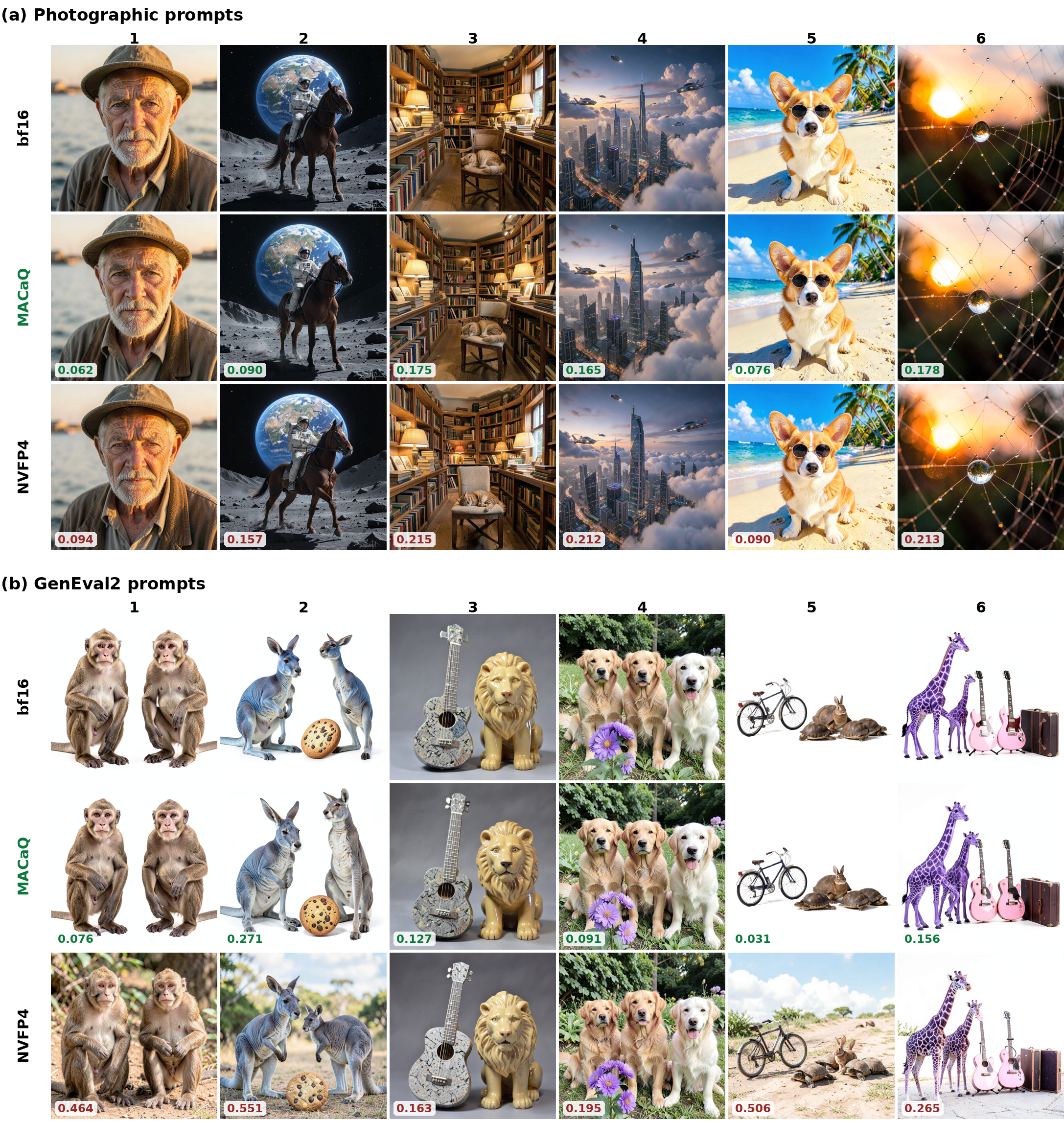}
    \caption{Qualitative comparison on FLUX.2-klein-4B: BF16 vs.\ MACaQ
(MXFP4 W4A4) vs.\ the official NVFP4 checkpoint, scaled to 512 pix width. Same seed per prompt. The inset number shows the lpips (lower is better). For a quantitative comparison see Table~\ref{tab:flux}.
\textbf{(a) Photographic prompts:}
(1) a photorealistic portrait of an elderly fisherman, weathered face, golden hour light;
(2) an astronaut riding a horse on the moon, earth in the background, dramatic;
(3) a cozy bookshop interior, warm lamps, stacks of old books, cat sleeping on a chair;
(4) a futuristic city skyline at dusk, flying cars, volumetric clouds, highly detailed;
(5) a corgi wearing tiny sunglasses on a tropical beach, vibrant colors;
(6) a macro photograph of a dewdrop on a spider web at sunrise.
\textbf{(b) GenEval2 prompts:}
(1) four brown monkeys;
(2) a blue kangaroo and a cookie;
(3) a stone guitar and a plastic lion;
(4) two purple flowers in front of three dogs;
(5) a bicycle to the left of a rabbit, and five brown turtles;
(6) three purple giraffes to the left of two pink guitars, and two suitcases.
Best viewed zoomed in.}
    \label{fig:Flux-appendix}
\end{figure*}
\section{Extending Flip-Refinement to Vector-Quantized Weights}
\label{app:vq}

Flip-Refinement can also refine vector-quantized weights by replacing a
codeword rather than a scalar code. We apply this extension to the
\texttt{IQ3\_S} and \texttt{IQ3\_XXS} weight-only formats in
\texttt{llama.cpp}.

\paragraph{Codebook constraints.}
Both formats encode four consecutive weight magnitudes with one codebook
index: \texttt{IQ3\_S} has $512$ codewords and \texttt{IQ3\_XXS} has $256$.
Each group of $32$ weights shares a 4-bit scale code $k$, and each group of
$256$ weights shares an FP16 scale $d$. A reconstructed weight is
$\widehat w_j=d\,f(k)\,\sigma_j c_j$, where $c$ is a codeword and
$\sigma_j\in\{\pm1\}$. The scale factors are $f(k)=1+2k$ for
\texttt{IQ3\_S} and $f(k)=(k+0.5)/2$ for \texttt{IQ3\_XXS}.
Signs are independent in \texttt{llama.cpp}'s \texttt{IQ3\_S} format;
\texttt{IQ3\_XXS} permits only the $128$ even-parity sign patterns within
each group of eight weights.

Changing one magnitude while holding the others fixed often produces a
tuple outside the codebook. Scalar Flip-Refinement therefore has few legal
moves, motivating a search over complete codewords.

\paragraph{Block refinement.}
We retain the regularized row objective from Sec.~\ref{sec:flip},
$L_i=r_i H_f r_i^\top$ with $r_i=w'_i-\widehat w'_i$ and
$g_i=H_f r_i^\top$. Recall that for a scalar move of size $\Delta$ on
input $j$, Eq.~\eqref{eq:flip-gain} gives the exact decrease
$2\Delta g_{i,j}-\Delta^2 (H_f)_{jj}$. A block move is the same statement
with vectors. Let $\mathcal{B}$ contain the four input indices of one
block, let $v$ and $u$ be its current and candidate reconstructions, and
write $A=(H_f)_{\mathcal{B},\mathcal{B}}$ and $\delta=u-v$. Then
\begin{equation}
  \operatorname{gain}(i,\mathcal{B},u)
  = 2\,\delta^\top g_{i,\mathcal{B}} - \delta^\top A\,\delta .
  \label{eq:vq-gain}
\end{equation}
The expression is exact because the objective is quadratic: a change
confined to four coordinates is described completely by the gradient on
those coordinates and the $4\times4$ Hessian block. Interactions with the
other $n-4$ weights of the row are already contained in
$g_{i,\mathcal{B}}$.

Eq.~\eqref{eq:vq-gain} has a geometric reading that makes the search
natural. Completing the square gives
\begin{equation}
\begin{aligned}
\operatorname{gain}(i,\mathcal{B},u)
&= \text{const}
 - (u-u_i^\star)^\top A\,(u-u_i^\star), \\
u_i^\star
&= v + A^{-1} g_{i,\mathcal{B}} .
\end{aligned}
\label{eq:vq-target}
\end{equation}
Here $u_i^\star$ is the value the four weights would take if they were
not quantized (the exact minimizer over the block with all other weights
fixed, i.e. one Newton step on the block), and the best codeword is the
one nearest to $u_i^\star$ in the metric defined by $A$. Block
refinement is therefore a nearest-neighbour search: for every row and
block, find the admissible codeword closest to that row's target. In the
scalar case of the main text this search has two candidates, the adjacent
levels, so its cost never mattered. Here it has up to $8192$ candidates,
and we organize it accordingly.

We score all admissible candidates and accept the best improving move.
For \texttt{IQ3\_S}, this means jointly searching magnitudes and signs over
$512\times16=8192$ signed codewords. For \texttt{IQ3\_XXS}, we alternate
between selecting one of $256$ codewords with signs fixed and selecting
one of $128$ legal sign patterns per eight-weight group with magnitudes
fixed. Each search is exact conditional on the quantities held fixed.

\paragraph{Efficient exact scoring.}
A nearest-neighbour search under a quadratic metric is evaluated the same
way as the assignment step of $k$-means: expand the square. Dropping the
terms of Eq.~\eqref{eq:vq-target} that do not depend on $u$ leaves
\begin{equation}
\begin{aligned}
S_i(u)
&= \underbrace{u^\top A\, u}_{\text{codeword ``length''}}
 - 2\,\underbrace{u^\top q_i}_{\text{match to row } i}, \\
q_i
&= A v + g_{i,\mathcal{B}} .
\end{aligned}
\label{eq:vq-split}
\end{equation}
and the codeword with the smallest $S_i$ is the nearest one. The two terms
have very different costs. The length term $u^\top A u$ depends on the
codeword and on the column block, but not on the row: the metric $A$ is a
block of $H_f$, which is shared by all $m$ output rows of the layer. It is
therefore a table with one entry per codeword, computed once per block and
reused by every row. The match term is a dot product between the row's
4-vector $q_i$ and the codeword. Computing it for all rows and all
candidates at once is a single matrix product, $Q\,\mathcal{C}^\top$, where
$Q\in\mathbb{R}^{m\times4}$ stacks the $q_i$ and
$\mathcal{C}\in\mathbb{R}^{C\times4}$ stacks the candidates. The score
matrix $(m\times C)$ is the table broadcast over rows minus twice this
product, and a row-wise argmin selects the codewords.

The block scale enters as a per-row scalar: a candidate is $u = s_i c$
with $c$ a signed codeword, so
$S_i(s_i c) = s_i^2\, c^\top A c - 2 s_i\, q_i^\top c$, and the table and
the product are combined with per-row factors. The table itself is
obtained with one small product per block: with
$\operatorname{vec}(cc^\top)\in\mathbb{R}^{16}$ precomputed for every
codeword, the $C$ lengths are $\operatorname{vec}(cc^\top)\cdot
\operatorname{vec}(A)$, a $(C\times16)(16)$ product.

For \texttt{IQ3\_S} the $16$ sign patterns of a block are folded into the
codebook, so one table of $C=8192$ lengths serves all rows. For
\texttt{IQ3\_XXS} the signs are held fixed during the magnitude search
and differ between rows, so the metric becomes row-dependent,
$(\sigma_i\sigma_i^\top)\odot A$: the table is now computed per row, as an
$(m\times16)(16\times256)$ product, which is still one GEMM. The sign
search over the $128$ parity-legal patterns of an 8-group has the same
structure with the magnitudes $\mu_i$ fixed and the metric
$(\mu_i\mu_i^\top)\odot (H_f)_{\mathcal{B},\mathcal{B}}$ on the
$8\times8$ block: an $(m\times64)(64\times128)$ product for the lengths
and an $(m\times8)(8\times128)$ product for the matches.

After a block is replaced in a set of rows, the gradients of those rows are
corrected exactly with the rank-4 analogue of Eq.~\eqref{eq:flip-update},
\begin{equation}
  g_i \leftarrow g_i - (H_f)_{:,\mathcal{B}}\,(u-v),
  \label{eq:vq-update}
\end{equation}
which adds four Hessian columns at a cost of $4n$ per changed row instead
of the $n^2$ of a fresh $H_f r_i^\top$. Only rows that accepted a move are
touched.

\paragraph{Sweeps and cost.}
Rows are processed in parallel, while blocks are visited sequentially in
random order using the updated gradients. We require a relative
improvement above $10^{-6}$ and recompute the gradients once per sweep to
limit numerical drift. Refinement stops after at most eight sweeps or when
a sweep accepts no move.

Per sweep over the $n/4$ blocks of an \texttt{IQ3\_S} tensor, the match
products amount to $2mnC$ operations in total, the length tables to
$O(nC)$, and the gradient updates to at most $O(mn^2)$ (if every row moved
at every block; in practice far fewer), the same order as the once-per-sweep
gradient refresh. The search thus scales linearly with the codebook size
and never re-evaluates a full row objective. For the down projection of
Qwen3-4B ($m=2560$, $n=9728$) with $C=8192$, the match products are
$4.1\times10^{11}$ operations per sweep. The inner dimension of these
products is only four and the blocks are processed sequentially, so peak
GPU throughput is not reached; eight sweeps of such a tensor, including
the sign and scale stages, take approximately $5$--$8$\,s. The FP32 score
matrix is the only sizeable intermediate ($84$\,MB at these dimensions);
for larger tensors we process rows in chunks to keep it below $600$\,MB.
All score products use FP32 with TF32 disabled, because the score
differences between neighbouring candidates can be small. Self-tests on
real GGUF tensors check monotone sweeps, agreement with brute-force block
minimization, and bit-exact repacking.

\paragraph{Optional scale refinement.}
With codes and signs fixed, we can also refine the shared scales. For a
group $\mathcal{S}$ reconstructed as $az$, define
$A_{\mathcal{S}}=(H_f)_{\mathcal{S},\mathcal{S}}$ and
$q=A_{\mathcal{S}}v+g_{i,\mathcal{S}}$. Up to a constant, the objective is
\begin{equation}
  L(a)=a^2 z^\top A_{\mathcal{S}}z-2a\, z^\top q .
\end{equation}
We evaluate all $16$ sub-block scale codes. For the super-block scale,
$z$ includes the sub-block factors, and we round the continuous minimizer
\begin{equation}
  d^\star=\frac{z^\top q}{z^\top A_{\mathcal{S}}z}
\end{equation}
to FP16. Scale changes are accepted only when they reduce the objective.
When enabled, these two scale stages follow the code and sign searches
in each sweep.

\paragraph{Setup.}
We quantize Qwen3-4B and Qwen3-14B with \texttt{llama.cpp} after applying
MACaQ's scaling and block-Hadamard rotation. The corresponding activation
transform runs online in a patched \texttt{llama.cpp}. Hessians are
estimated from $10^6$ WikiText-2 training tokens, transformed into the
quantization basis, and regularized with $\lambda=1$. The output head uses
the Hessian of the final-normalization output.

We compare the initial quantization, scalar flips restricted to legal
on-grid moves with $K=512$, and eight sweeps of block refinement. Initial
files are quantized either with or without \texttt{llama.cpp}'s importance
matrix, computed from the same calibration text. Perplexity is measured
with \texttt{llama-perplexity} on WikiText-2 and a Pile sample; these
measurements are not directly comparable to Tab.~\ref{tab:ppl}.

\begin{table}[t]
\centering
\caption{Weight-only quantization of MACaQ-transformed Qwen3 models.
Lower perplexity is better. \emph{imatrix} denotes initialization with
\texttt{llama.cpp}'s importance matrix; \emph{plain} omits it.
\emph{codes} refines codewords and signs only, while
\emph{codes+scales} additionally refines the shared scales.
Codes-only refinement was not run at 14B.}
\label{tab:vq}
\small

\begin{tabular}{llcc}
\toprule
Format & Refinement & Wiki2 & Pile \\
\midrule

\multicolumn{4}{l}{\textbf{Qwen3-4B}} \\
\midrule
F16
  & -- & 11.56 & 7.41 \\

\texttt{IQ3\_S} (plain)
  & none         & 13.06 & 8.10 \\
  & scalar flips & 12.31 & 7.85 \\
  & block, codes & \textbf{11.81} & \textbf{7.67} \\
  & block, codes+scales & 11.86 & 7.68 \\

\addlinespace
\texttt{IQ3\_S} (imatrix)
  & none         & 11.83 & 7.73 \\
  & scalar flips & 11.91 & 7.79 \\
  & block, codes & 11.88 & \textbf{7.72} \\
  & block, codes+scales & 11.89 & 7.73 \\

\addlinespace
\texttt{IQ3\_XXS} (imatrix)
  & none         & 13.09 & 8.36 \\
  & scalar flips & 12.76 & 8.21 \\
  & block, codes & \textbf{12.64} & \textbf{8.14} \\
  & block, codes+scales & 12.82 & 8.21 \\

\midrule
\multicolumn{4}{l}{\textbf{Qwen3-14B}} \\
\midrule
F16
  & -- & 7.52 & 5.63 \\

\texttt{IQ3\_S} (plain)
  & none         & 7.75 & 5.79 \\
  & scalar flips & 7.74 & 5.80 \\
  & block, codes+scales & \textbf{7.68} & \textbf{5.75} \\

\addlinespace
\texttt{IQ3\_S} (imatrix)
  & none         & 7.69 & 5.73 \\
  & scalar flips & \textbf{7.61} & \textbf{5.71} \\
  & block, codes+scales & 7.64 & 5.73 \\

\addlinespace
\texttt{IQ3\_XXS} (imatrix)
  & none         & 8.07 & 5.96 \\
  & scalar flips & \textbf{7.92} & \textbf{5.90} \\
  & block, codes+scales & 7.98 & 5.95 \\

\bottomrule
\end{tabular}
\end{table}
\paragraph{Results.}
Block refinement is most effective without an importance matrix
(Tab.~\ref{tab:vq}). On Qwen3-4B in plain \texttt{IQ3\_S}, it reduces
WikiText-2 perplexity from $13.06$ to $11.81$, compared with $12.31$ for
scalar flips and $11.83$ for importance-matrix initialization alone.
On Qwen3-4B, block refinement reduces individual tensors' Hessian objectives
by $30$--$42\%$, versus $16$--$32\%$ for scalar flips, and takes approximately
$5$--$8$ seconds per large tensor.

With importance-matrix initialization, the gains depend on the format.
Refinement changes \texttt{IQ3\_S} perplexity only slightly, but improves
\texttt{IQ3\_XXS} on both models. Among the refinements, block search
performs best at $4$B, while scalar flips perform best at $14$B. Exact
block minimization thus does not guarantee better model perplexity: the
searches can reach different local minima, and the layer objective is only
a proxy for model quality.

Scale refinement further reduces the calibration objective but does not
improve perplexity over codes-only refinement in the $4$B comparisons.
We therefore recommend codes-only block refinement for these formats.

\end{document}